\documentclass[runningheads]{llncs}
\usepackage{graphicx}
\usepackage{booktabs}
\usepackage[switch]{lineno}
\usepackage{amsmath}
\usepackage{amsfonts}
\usepackage{amssymb}

\begin{document}
\title{CaSS: A Channel-aware Self-supervised Representation Learning Framework for Multivariate Time Series Classification}
\titlerunning{CaSS: A Channel-aware Self-supervised Framework}

\author{Yijiang Chen\inst{1} \and
Xiangdong Zhou\inst{1} \and
Zhen Xing\inst{1} \and Zhidan Liu\inst{1} \and Minyang Xu\inst{1,2}}
\authorrunning{Y. Chen et al.}
\institute{School of Computer Science, Fudan University, Shanghai, China, 200433\\ \email{\{chenyj20,xdzhou,zxing20,zdliu20,16110240027\}@fudan.edu.cn} \and Arcplus Group PLC, Shanghai, China, 200041}

\maketitle              % typeset the header of the contribution
\begin{abstract}
Self-supervised representation learning of Multivariate Time Series (MTS) is a challenging task and attracts increasing research interests in recent years. Many previous works focus on the pretext task of self-supervised learning and usually neglect the complex problem of MTS encoding, leading to unpromising results. In this paper, we tackle this challenge from two aspects: encoder and pretext task, and propose a unified channel-aware self-supervised learning framework CaSS. Specifically, we first design a new Transformer-based encoder Channel-aware Transformer (CaT) to capture the complex relationships between different time channels of MTS. Second, we combine two novel pretext tasks Next Trend Prediction (NTP) and Contextual Similarity (CS) for the self-supervised representation learning with our proposed encoder. Extensive experiments are conducted on several commonly used benchmark datasets. The experimental results show that our framework achieves new state-of-the-art comparing with previous self-supervised MTS representation learning methods (up to +7.70\% improvement on LSST dataset) and can be well applied to the downstream MTS classification.
\end{abstract}
\section{Introduction}
With the fast progress of IoT and coming 5Gs, multivariate time series  widely exists in medical, financial, industrial and other fields as an increasingly important data form \cite{r39,r40}. Compared with univariate time series, multivariate time series usually contains more information and brings more potentials for data mining, knowledge discovery and decision making, etc.. However, MTS not only contains time-wise patterns, but also has complex relationships between different channels, which makes MTS analysis much more difficult. 

In recent years, the self-supervised learning attracts more and more attention from research and industry communities. The self-supervised pre-training demonstrates its success in the fields of Natural Language Processing (NLP) \cite{r27} and Computer Visions (CV) \cite{r41}. Especially in NLP, adopting a pre-trained language model is de facto the first step of almost all the NLP tasks. Likewise, the self-supervised representation learning of time series not only brings performance improvements, but also helps to close the gap between the increasing amount of data (more abstraction and complexity)  and the expensive cost of manually labeling for supervised learning tasks.  Many research efforts have been devoted to the self-supervised representation learning of time series \cite{r22,r23,r26} and promising results have been achieved. However, the previous works for MTS are limited and the challenge still exists.

\begin{figure}[t]
\centering
\includegraphics[width=1\textwidth]{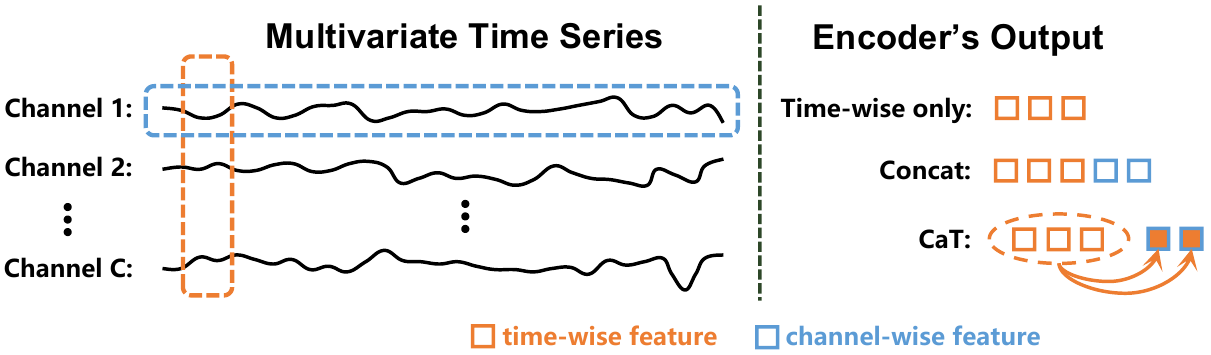} % Reduce the figure size so that it is slightly narrower than the column.
\caption{A sample of encoding of multivariate time series.}
\label{fig0}
\end{figure}

The self-supervised representation learning usually consists of two aspects: encoder and pretext task. As shown in Figure \ref{fig0}, in the past works, most of them only focus on time-wise features where all channel values of one or several time steps are fused through convolution or fully connected layer directly in the embedding process. There is a lack of deliberate investigation of the relationships between channel-wise features, which affects the encoder’s ability to capture the whole characteristics of the MTS. To deal with the problem, the methods 
combining with Recurrent Neural Network (RNN) are presented to capture the individual feature of each channel \cite{r14,r36}. The recent work of \cite{r21} employs Transformer \cite{r29} to integrate the features of time-wise and channel-wise. Among these solutions, RNN seems not very suitable for self-supervised leaning due to the consumption and is usually employed in prediction task \cite{r11}. Transformer is becoming more and more popular and is suitable for time series \cite{r26}, however the previous Transformer-based MTS embedding provokes the problem of high complexity of computing and space, which prevents it from real applications. It inspires us to design a more effective Transformer-based for MTS to take advantages of the strong encoding ability of Transformer. In the aspect of pretext task, most of the previous works adopt the traditional time series embedding based on time-wise features. How to integrate channel-wise features with pretext task is a challenge.

In this paper, we propose a novel self-supervised learning framework CaSS from the aspects of encoder and pretext task. First, we propose a new Transformer-based encoder Channel-aware Transformer (CaT) for MTS encoding which investigates time-wise and channel-wise features simultaneously. It is noticed that in practice the number of channels of MTS is fixed while the time length can be unlimited and the number of channels is usually much less than the time length. Therefore as shown in Figure \ref{fig0}, different from previous work \cite{r21}, we integrate the time-wise features into the channel-wise features and concatenate all these novel channel-wise features as the representation of the sample. Second, we design a new self-supervised pretext task Next Trend Prediction (NTP) from the perspective of channel-wise for the first time in self-supervised MTS representation learning. It is considered that in many cases only the rise and fall of future time rather than the specific value of the time series is necessary. So we cut the multivariate time series from the middle, and using the previous sequences of all rest channels to predict the trend for each channel. Different from fitting the specific value (regression), the prediction of trend (rise and fall) is more suitable for arbitrary data. We also demonstrate through experiments that compared with fitting specific values, prediction of trend is more efficient. In addition, we employ another task called Contextual Similarity (CS) which combines a novel data augmentation strategy to maximize the similarity between similar samples and learn together with NTP task. The CS task focuses on the difference between samples while NTP task focuses on the sample itself and helps to learn the complex internal characteristics.

In summary, the main contributions of our work are as follows:

\begin{itemize}
\item We propose a new Transformer-based encoder Channel-aware Transformer. It can efficiently integrate the time-wise features to the channel-wise representation.
\item We design two novel pretext tasks, Next Trend Prediction and Contextual Similarity for our CaSS framework. To the best of our knowledge, Next Trend Prediction task is conducted from the perspective of channel-wise for the first time in self-supervised MTS representation learning.
\item We conduct extensive experiments on several commonly used benchmark datasets from different fields. Compared with the state-of-the-art self-supervised MTS representation learning methods, our method achieves new state-of-the-art in MTS classification (up to +7.70\% improvement on LSST dataset). We also demonstrate its ability in few-shot learning.

\end{itemize}

\section{Related Work}

\subsection{Encoders for Time Series Classification} 
A variety of methods have been proposed for time series classification. Early works employ traditional machine learning methods to solve the problem, like combining Dynamic Time Warping (DTW) \cite{r1} with Support Vector Machine (SVM) \cite{r2}. Time Series Forest \cite{r3} introduces an approach based on Random Forest \cite{r42}. Bag of Patterns (BOP) \cite{r4} and Bag of SFA Symbols (BOSS) \cite{r5} construct a dictionary-based classifier. Although these early works can deal with the problem to some extent, they need heavy crafting on data preprocessing and feature engineering. The emergence of deep learning greatly reduces feature engineering and boosts the performance of many machine learning tasks. So far Convolutional Neural Network (CNN) is popular in time series classification due to its balance between its effect and cost, such as Multi-scale Convolutional Neural Network (MCNN) \cite{r6} for univariate time series classification, Multi-channels Deep Convolutional Neural Networks (MC-DCNN) \cite{r7} for multivariate time series classification and so on \cite{r31,r32,r33}. Hierarchical Attention-based Temporal Convolutional Network (HA-TCN) \cite{r9} and WaveATTentionNet (WATTNet) \cite{r37} apply dilated causal convolution to improve the encoder’s effect. Among these CNN methods, Fully Convolutional Network (FCN) and Residual Network (ResNet) \cite{r10} have been proved to be the most powerful encoders in multivariate time series classification task \cite{r11}. Due to the high computation complexity, RNN based encoders are rarely applied solely to the time series classification \cite{r35,r36}. It is often combined with CNN to form a two tower structure \cite{r12,r14}. In recent years, more and more works have tried to apply Transformer to time series \cite{r15,r16,r17}. However, most of them are designed for prediction task, and few works cover the classification problem \cite{r19,r20,r21,r26}.

\subsection{Pretext Tasks for Time Series}
Manually labeling is a long lasting challenge for the supervised learning, and recently self-supervised training (no manually labeling) becomes more and more popular in many research fields including time series analysis. To name a few, \cite{r22} employs the idea of word2vec \cite{r28} which regards part of the time series as word, the rest as context, and part of other time series as negative samples for training. \cite{r23} employs the idea of contrastive learning where two positive samples are generated by weak augmentation and strong augmentation to predict each other while the similarity among different augmentations of the same sample is maximized. \cite{r24} is designed for univariate time series. It samples several segments of the time series and labels each segment pairs according to their relative distance in the origin series. It also adds the task of judging whether two segments are generated by the same sample. \cite{r25} is based on sampling pairs of time windows and predicting whether time windows are close in time by setting thresholds to learn EEG features. \cite{r26} is a Transformer-based method which employs the idea of mask language model \cite{r27}. The mask operation is performed for multivariate time series and the encoder is trained by predicting the masked value. However, the previous works usually focus on time-wise features and need to continuously obtain the features of several time steps \cite{r22,r23,r26}, which makes them difficult to be applied to the novel MTS representation.

\begin{figure*}[t]
\centering
\includegraphics[width=1\textwidth]{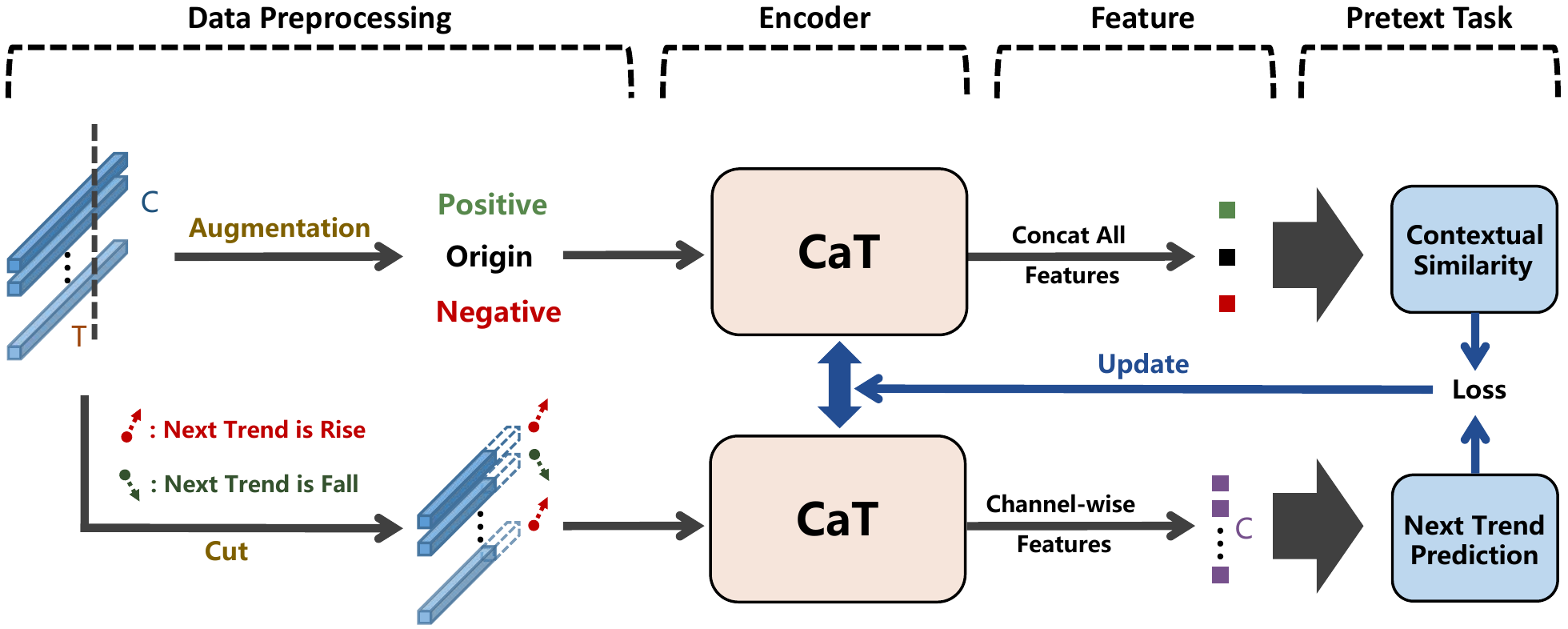} % Reduce the figure size so that it is slightly narrower than the column.
\caption{Overall architecture of the channel-aware self-supervised framework CaSS.}
\label{fig1}
\end{figure*}

\section{The Framework}
In this work, we focus on self-supervised multivariate time series representation learning. Given $M$ multivariate time series $X$= \{$x_0$,...,$x_M$\}, where $x_i\in \mathbb{R}^{C \times T}$ refers to the $i$-th time series which has $C$ channels and $T$ time steps. For each multivariate time series $x_i$, our goal is to generate a proper representation $z_i$ which is applicable to the subsequent task.

Our proposed novel encoder Channel-aware Transformer (CaT) and pretext tasks constitute our channel-aware self-supervised learning framework CaSS. The overall architecture is shown in Figure \ref{fig1}. In our framework, CaT is to generate the novel channel-wise features of the MTS sample, and the generated features are served as the inputs of Next Trend Prediction task and Contextual Similarity task. Specifically, we first preprocess the time series samples and apply the two pretext tasks to learn the encoder (representations), then we employ the learnt representations to the MTS classification task by freezing the encoder.

\section{Channel-aware Transformer}
\begin{figure*}[t]
\centering
\includegraphics[width=1\textwidth]{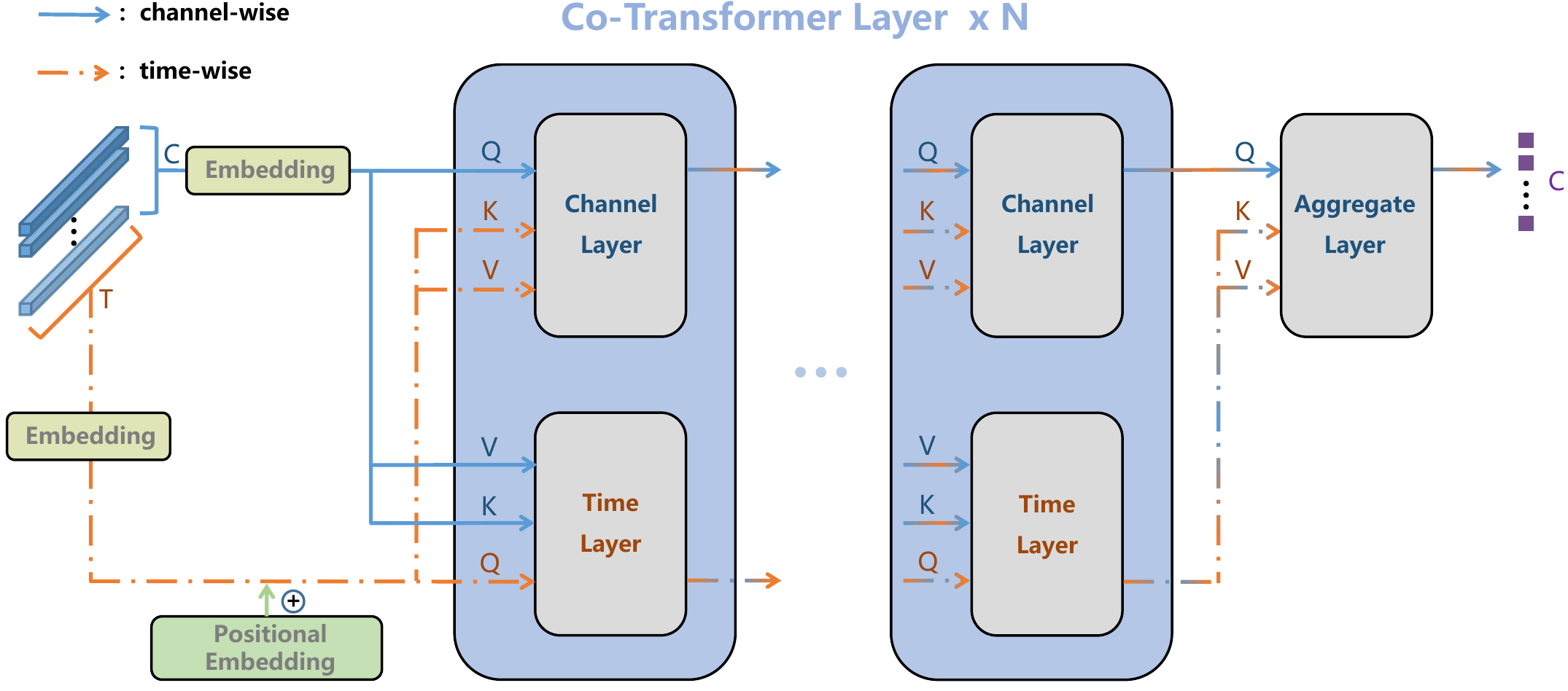} % Reduce the figure size so that it is slightly narrower than the column.
\caption{Overall architecture of Channel-aware Transformer (CaT). Q, K, V represent the query vector, key vector, value vector in attention operation respectively.}
\label{fig2}
\end{figure*}

This section describes our proposed Transformer-based encoder Channel-aware Transformer which is served as the encoder in our self-supervised learning framework. As shown in Figure \ref{fig2}, It consists of Embedding Layer, Co-Transformer Layer and Aggregate Layer. The two Transformer structures in Co-Transformer Layer encode the time-wise and channel-wise features respectively, and interact with each other during encoding. Finally, we fuse the time-wise features into channel-wise features through Aggregate Layer to generate the final representation.

\subsection{Embedding Layer}
Given an input sample $x\in \mathbb{R}^{C \times T}$, we map it to the $D$-dimension time vector space and channel vector space respectively to obtain the time embedding $e_t\in \mathbb{R}^{T \times D}$ and channel embedding $e_c\in \mathbb{R}^{C \times D}$:

\begin{equation}
    e_{t}=x^{T} W_{t}+e_{pos},
\end{equation}

\begin{equation}
e_{c}=x W_{c},
\end{equation}
where $W_t\in \mathbb{R}^{C \times D}$, $W_c\in \mathbb{R}^{T \times D}$ are learnable embedding matrices. $e_{pos}\in \mathbb{R}^{T \times D}$ is the positional embedding applying the design of \cite{r29}.

\subsection{Co-Transformer Layer}
This part adopts a $N$-layer two tower structure based on Transformer \cite{r21}. Each layer is composed of Time Layer and Channel Layer, focusing on time-wise and channel-wise respectively. Supposing in the $i$-th layer $(i=0,...,N-1)$, we obtain the input $a_t^{i} \in \mathbb{R}^{T \times D}$ for Time Layer and $a_c^{i} \in \mathbb{R}^{C \times D}$ for Channel Layer. The process in Time Layer is:

\begin{equation}
    Q_{t}^{i}=a_t^{i}W_{qt}^{i}, \quad
    K_{t}^{i}=a_c^{i}W_{kt}^{i}, \quad
    V_{t}^{i}=a_c^{i}W_{vt}^{i}, \quad
\end{equation}

\begin{equation}
    b_{t}^{i}=\text{LayerNorm}(\text{MHA}(Q_{t}^{i}, K_{t}^{i}, V_{t}^{i})+a_t^{i}),
\end{equation}

\begin{equation}
    a_{t}^{i+1}=\text{LayerNorm}(\text{FFN}(b_t^{i})+b_t^{i}),
\end{equation}
and the process in Channel Layer is:

\begin{equation}
    Q_{c}^{i}=a_c^{i}W_{qc}^{i}, \quad
    K_{c}^{i}=a_t^{i}W_{kc}^{i}, \quad
    V_{c}^{i}=a_t^{i}W_{vc}^{i}, \quad
\end{equation}

\begin{equation}
    b_{c}^{i}=\text{LayerNorm}(\text{MHA}(Q_{c}^{i}, K_{c}^{i}, V_{c}^{i})+a_c^{i}),
\end{equation}

\begin{equation}
    a_{c}^{i+1}=\text{LayerNorm}(\text{FFN}(b_c^{i})+b_c^{i}),
\end{equation}
where $W_{qt}^{i}$, $W_{kt}^{i}$, $W_{vt}^{i}$, $W_{qc}^{i}$, $W_{kc}^{i}$, $W_{vc}^{i}$ $\in \mathbb{R}^{D \times D}$ are learnable matrices. MHA is the abbreviation of multi-head attention and FFN is the abbreviation of feed forward network. Specifically, $a_t^{0}=e_t$, $a_c^{0}=e_c$.

This interactive way helps reduce the time complexity from $o((T^2+C^2)D)$ to $o(2TCD)$ compared with the non-interactive two tower Transformer-based encoder which applies self-attention mechanism \cite{r29}. In the real world, $T$ is usually much larger than $C$, so it can further boost the speed of the encoder.

\subsection{Aggregate Layer}
Through Co-Transformer Layer, we can obtain time-wise features $a_{t}^{N}$ and channel-wise features $a_{c}^{N}$. If all features are forcibly concatenated, the final dimension of the representation will be too large to be applied in the subsequent work. In real applications, the channel length is usually much less than the time length, therefore we integrate the time-wise features into the channel-wise features through attention operation. Finally, we concatenate these novel channel-wise features as the final representation $z \in \mathbb{R}^{1 \times (C\cdot D)}$:

\begin{equation}
    Q_{c}^{N}=a_c^{N}W_{qc}^{N}, \quad
    K_{c}^{N}=a_t^{N}W_{kc}^{N}, \quad
    V_{c}^{N}=a_t^{N}W_{vc}^{N}, \quad
\end{equation}

\begin{equation}
    a_{c}=\text{MHA}(Q_{c}^{N}, K_{c}^{N}, V_{c}^{N}),
\end{equation}

\begin{equation}
    z=[a_c^{1}, a_c^2, ..., a_c^C],
\end{equation}
where $W_{qc}^{N}, W_{kc}^{N}, W_{vc}^{N} \in \mathbb{R}^{D \times D}$ are learnable matrices. [·,·] is the concatenation operation.

\section{Pretext Task}
In order to enable our encoder to carry out self-supervised learning more efficiently, we design two novel pretext tasks Next Trend Prediction (NTP) and Contextual Similarity (CS) based on our novel channel-wise representation.

\subsection{Next Trend Prediction}
Given a sample $x_{i}\in \mathbb{R}^{C \times T}$, we randomly select a time point $t\in [1, T-1]$ for truncation. The sequence before $t$ time step $x_{i}^{NTP(t)}\in \mathbb{R}^{C\times t}$ is regarded as the input of the NTP task and the data after $t$ is padded to $T$ with zeros. For each channel $j$, we adopt the trend of the $t+1$ time step as the label $y_{i,j}^{NTP(t)}$ for training:

\begin{equation}
y_{i,j}^{NTP(t)}= \left \{
\begin{array}{ll}
    1,     & \text{if}\quad x_{i}[j, t+1]\geq x_{i}[j, t] \\
    0,     & \text{if}\quad x_{i}[j, t+1]< x_{i}[j, t] \\
\end{array},
\right.
\end{equation}
where $x_{i}[j, t]$ represents the value of $t$ time step of channel $j$ in $x_{i}$.

After inputting the NTP sample $x_{i}$ into our encoder, we can obtain the representation $z_{i}^{NTP(t)}\in \mathbb{R}^{C\times D}$ where $z_{i,j}^{NTP(t)}\in \mathbb{R}^{D}$ represents the representation of the $j$-th channel. Finally, a projection head is applied to predict the probability of the rise and fall. Assuming that every sample generated $K_{NTP}$ input samples and the corresponding truncating time point set is $S\in \mathbb{R}^{K_{NTP}}$. For $x_{i}$, the loss of the NTP task can be obtained through the following formula:

\begin{equation}
\ell_{NTP}=\sum\nolimits_{t\in S}\sum_{j}^{C} \text{CE}(\varphi_{0}(z_{i,j}^{NTP(t)}), y_{i,j}^{NTP(t)}),
\end{equation}
where $\varphi_{0}$ is the projection head of the NTP task and CE is the Cross Entropy loss function.

\subsection{Contextual Similarity}
The purpose of NTP task is to enable the sample to learn the relationships between its internal channels. Meanwhile, we need to ensure the independence between different samples, so we further employ the Contextual Similarity task. The main difference between CS task and Contextual Contrasting task in TS-TCC \cite{r23} lies in the design of the augmentation method. To help the framework focus more on the dependencies between channels, we further apply asynchronous permutation strategy to generate negative samples. And we also add the original sample to the self-supervised training to enhance the learning ability.

In this task, we generate several positive samples and negative samples for each sample through augmentation strategy. For positive samples, we adopt the Interval Adjustment Strategy which randomly selects a series of intervals, and then adjust all values by jittering. Further, we also adopt a Synchronous Permutation Strategy which segments the whole time series and disrupts the segment order. It helps to maintain the relations between segments and the generated samples are regarded as positive. For negative samples, we adopt an Asynchronous Permutation Strategy which randomly segments and disrupts the segment order for each channel in different ways. In experiments, for each sample, we use the interval adjustment strategy and the synchronous disorder strategy to generate one positive sample respectively, and we use the asynchronous disorder strategy to generate two negative samples. Assuming that the batch size is $B$, we can generate extra $4B$ augmented samples. Therefore the total number of the sample in a batch is $5B$.

With the $i$-th sample $x_{i}$ in a batch our encoder can obtain its representation $z_{i}^{0}\in \mathbb{R}^{1\times(C\cdot D)}$. The representations of inputs except itself are $z_{i}^{*}\in \mathbb{R}^{(5B-1)\times (C\cdot D)}$, where $z_{i}^{*,m}\in \mathbb{R}^{1\times (C\cdot D)}$ is the $m$-th representation of $z_{i}^{*}$. Among $z_{i}^{*}$, the two positive samples are $z_{i}^{+,1},z_i^{+,2}\in \mathbb{R}^{1\times(C\cdot D)}$. Therefore, for $i$-th sample, the loss of the CS task can be obtained through the following formula:

\begin{equation}
    \ell_{CS}\!=\!-\!\sum_{n=1}^{2}\! \text{log}\frac{ \text{exp}(\text{sim}(\varphi_{1}(z_{i}^{0}), \varphi_{1}(z_{i}^{+,n}))/\tau)}{\sum_{m=1}^{5B-1}\!\text{exp}(\text{sim}(\varphi_{1}(z_{i}^{0}), \varphi_{1}(z_{i}^{*,m}))/\tau) },
\end{equation}
where $\tau$ is a hyperparameter, $\varphi_{1}$ is the projection head of the CS task, sim is the cosine similarity.

The final self-supervised loss is the combination of the NTP loss and CS loss as follows:

\begin{equation}
    \ell = \alpha_{1}\cdot\ell_{NTP}+\alpha_{2}\cdot\ell_{CS},
\end{equation}
where $\alpha_{1}$ and $\alpha_{2}$ are hyperparameters.

\section{Experiments}
\subsection{Datasets}
To demonstrate the effectiveness of our self-supervised framework, we use the following four datasets from different fields:

\begin{itemize}
\item \textbf{UCI HAR} \footnote{\url{https://archive.ics.uci.edu/ml/datasets/Human+Activity+Recognition+Using+Smartphones}}: The UCI HAR dataset \cite{r43} contains sensor readings of 6 human activity types. They are collected with a sampling rate of 50 Hz.
\item \textbf{LSST} \footnote{\url{http://www.timeseriesclassification.com/description.php?Dataset=LSST}}: The LSST dataset \cite{r44} is an open data to classify simulated astronomical time series data in preparation for observations from the Large Synoptic Survey Telescope.
\item \textbf{ArabicDigits} \footnote{\url{http://www.mustafabaydogan.com/}\label{web3}}: The ArabicDigits dataset \cite{r34} contains time series of mel-frequency cepstrum coefficients corresponding to spoken Arabic digits. It includes data from 44 male and 44 female native Arabic speakers.
\item \textbf{JapaneseVowels} \textsuperscript{\ref{web3}}: In the JapaneseVowels dataset \cite{r34}, several Japanese male speakers are recorded saying the vowels ‘a’ and ‘e’. A ‘12-degree linear prediction analysis’ is applied to the raw recordings to obtain time-series with 12 dimensions.
\end{itemize}
The detailed information is shown on Table \ref{t1}.

\begin{table}[htbp]
\centering
\caption{Detailed information of the used datasets.}
\scalebox{1}{
\begin{tabular}{lccccc}
\specialrule{0em}{0.5pt}{0.5pt}
\toprule
\textbf{Dataset} & \textbf{Train} & \textbf{Test} & \textbf{Time} & \textbf{Channel} & \textbf{Class} \\   \midrule

\textbf{UCI HAR} & 7352 & 2947 & 128 & 9 & 6  \\  
\textbf{LSST} & 2459 & 2466 & 36 & 6 & 14  \\  
\textbf{ArabicDigits} & 6600 & 2200 & 93 & 13 & 10  \\  
\textbf{JapaneseVowels} & 270 & 370 & 29 & 12 & 9  \\  

\bottomrule
\end{tabular}}
\label{t1}
\end{table}

\subsection{Experimental Settings}

% \begin{itemize}
% \item For supervised learning, we set $D$ = 512, $N$ = 8, attention head = 8, batch size = 4, dropout = 0.2. We use Adam optimizer with a learning rate of 1e-4.
% \item For self-supervised learning, we set $K_{NTP}$ = 10, $D$ = 512, $N$ = 8, attention head = 8, batch size = 10, dropout = 0.2, $\tau$ = 0.2, $\alpha_{1}$ = 2, $\alpha_{2}$ = 1. We use Adam optimizer with a learning rate of 5e-5.
% \item For fine-tuning after self-supervised learning, we train a single fully connected layer on top of the frozen self-supervised pre-trained encoder to evaluate the effect of our self-supervised framework. We set batch size = 4 and use Adam optimizer with a learning rate of 1e-3.
% \item We evaluate the performance using two metrics: Accuracy (ACC) and Macro-F1 score (MF1). Every result is generated by repeating 5 times with 5 different seeds.
% \item We conduct our experiments using PyTorch 1.7 and train models on a NVIDIA GeForce RTX 2080 Ti GPU.
% \end{itemize}

For supervised learning, we set $D$ = 512, $N$ = 8, attention head = 8, batch size = 4, dropout = 0.2. We use Adam optimizer with a learning rate of 1e-4.

For self-supervised learning, we set $K_{NTP}$ = 10, $D$ = 512, $N$ = 8, attention head = 8, batch size = 10, dropout = 0.2, $\tau$ = 0.2, $\alpha_{1}$ = 2, $\alpha_{2}$ = 1. We use Adam optimizer with a learning rate of 5e-5.

For fine-tuning after self-supervised learning, we train a single fully connected layer on top of the frozen self-supervised pre-trained encoder to evaluate the effect of our self-supervised framework. We set batch size = 4 and use Adam optimizer with a learning rate of 1e-3.

We evaluate the performance using two metrics: Accuracy (ACC) and Macro-F1 score (MF1). Every result is generated by repeating 5 times with 5 different seeds.

We conduct our experiments using PyTorch 1.7 and train models on a NVIDIA GeForce RTX 2080 Ti GPU.

\subsection{Baselines}
We compare our framework against the following self-supervised methods. It is noted that we apply the default hyperparameters of each compared method from the original paper or the code. The detailed information and the reason why we choose these methods are as followed: \\
\textbf{(1) W2V} \cite{r22}: This method employs the idea of word2vec. It combines an encoder based on causal dilated convolutions with a triplet loss and time-based negative sampling. Finally, they train a SVM on top of the frozen self-supervised pre-trained encoder. It achieves great results leveraging unsupervised learning for univariate and multivariate classification datasets. We select $K=10, 20$ from the original experiments. \\ 
\textbf{(2) W2V+}: It applies our proposed two tower Transformer-based model as the encoder while training with the pretext task of W2V. For the requirement of the time-wise feature, we replace Channel-aware Transformer with Time-aware Transformer (TaT) which integrates the channel-wise features into the time-wise features in the Aggregate Layer. \\ 
\textbf{(3) TS-TCC} \cite{r23}: This method employs the idea of contrastive learning using a convolutional architecture as encoder. After self-supervised learning, they train a fully connected layer on top of the frozen self-supervised pre-trained encoder. It is the state-of-the-art contrastive learning method in the field of self-supervised of time series. \\ 
\textbf{(4) TS-TCC+}: It applies TaT as the encoder like W2V+ while training with the pretext task of TS-TCC. \\ 
\textbf{(5) TST} \cite{r26}: This method employs the idea of masked language model using a Transformer-based encoder. In their work, it achieves outstanding results by finetuning the whole encoder after pre-training it. For fair comparison, we freeze the encoder while finetuning. \\ 
\textbf{(6) TST+}: It applies TaT as the encoder like W2V+ while training with the pretext task of TST. \\ 
\textbf{(7) NVP+CS}: To compare with the regression, in our framework we replace Next Trend Prediction with Next Value Predict (NVP) and regard it as a new strong baseline. We select 15\% time steps to predict the values of the next time step. \\ 
\textbf{(8) Supervised}: Supervised learning on both encoder and fully connected layer. \\

\begin{table*}
\centering
\caption{Comparison between CaSS and other self-supervised methods. ${\uparrow}$ mark indicates that the self-supervised result performs better than the supervised result.}
\scalebox{0.68}{
\begin{tabular}{lcccccccc}
\specialrule{0em}{0.5pt}{0.5pt}
\toprule
\centering
& \multicolumn{2}{c}{\textbf{HAR}} & \multicolumn{2}{c}{\textbf{LSST}} & \multicolumn{2}{c}{\textbf{ArabicDigits}} & \multicolumn{2}{c}{\textbf{JapaneseVowels}} \\ \midrule
\textbf{Method} & \textbf{ACC} & \textbf{MF1} & \textbf{ACC} & \textbf{MF1} & \textbf{ACC} & \textbf{MF1} & \textbf{ACC} & \textbf{MF1}\\   \midrule

\textbf{W2V K=10} & 90.37$\pm$0.34 & 90.67$\pm$0.97 & 57.24$\pm$0.24 & 36.97$\pm$0.49 & 90.16$\pm$0.57 & 90.22$\pm$0.52 & 97.98$\pm$0.40 & 97.73$\pm$0.48  \\
\textbf{W2V K=20} & 90.08$\pm$0.18 & 90.10$\pm$0.16 & 53.47$\pm$0.75 & 32.84$\pm$1.25 & 90.55$\pm$0.77 & 90.60$\pm$0.75 & 97.98$\pm$0.40 & 97.89$\pm$0.43  \\
\textbf{W2V+} & 84.55$\pm$0.59 & 84.34$\pm$0.69 & 54.90$\pm$0.58 & 33.97$\pm$0.79 & 87.59$\pm$0.48 & 87.56$\pm$0.49 & 95.27$\pm$0.38 & 95.25$\pm$0.39  \\
\textbf{TS-TCC} & 90.74$\pm$0.25 & 90.23$\pm$0.29 & 40.38$\pm$0.35 & 23.93$\pm$1.93 & 95.64$\pm$0.37 & 95.43$\pm$0.37 & 82.25$\pm$1.16 & 82.04$\pm$1.17 \\
\textbf{TS-TCC+} & 90.87$\pm$0.31 & 90.86$\pm$0.30 & 50.75$\pm$0.23 & 32.87$\pm$0.65 & 96.87$\pm$0.34 & 96.80$\pm$0.28 & 84.36$\pm$0.27 & 84.09$\pm$0.21 \\
\textbf{TST} & 77.62$\pm$2.48 & 78.05$\pm$2.56 & 32.89$\pm$0.04 & 7.86$\pm$1.63 & 90.73$\pm$0.36 & 90.90$\pm$0.33 & 97.30$\pm$0.27 & 97.47$\pm$0.34  \\
\textbf{TST+} & 87.39$\pm$0.49 & 87.77$\pm$0.11 & 34.49$\pm$0.38 & 14.62$\pm$0.32 & 96.82$\pm$0.23 & 96.82$\pm$0.22 & 97.87$\pm$0.23$^{\uparrow}$ & 97.51$\pm$0.22  \\
\textbf{NVP+CS} & 92.47$\pm$0.20 & 92.38$\pm$0.19 & 32.42$\pm$0.14 & 6.02$\pm$0.30 & 96.50$\pm$0.45 & 96.51$\pm$0.55 & 96.34$\pm$0.43 & 95.49$\pm$0.11  \\
\textbf{CaSS} & \textbf{92.57$\pm$0.24$^{\uparrow}$} & \textbf{92.40$\pm$0.17$^{\uparrow}$} & \textbf{64.94$\pm$0.02} & \textbf{46.11$\pm$0.55} & \textbf{97.07$\pm$0.20} & \textbf{97.07$\pm$0.20} & \textbf{98.11$\pm$0.27$^{\uparrow}$} & \textbf{98.14$\pm$0.08$^{\uparrow}$}  \\ \midrule

\textbf{Supervised} & 92.35$\pm$0.63 & 92.40$\pm$0.58 & 66.57$\pm$0.38 & 51.60$\pm$1.26 & 98.07$\pm$0.38 & 98.07$\pm$0.38 & 97.71$\pm$0.13 & 97.56$\pm$0.07  \\

\bottomrule
\end{tabular}}
\label{t2}
\end{table*}

\subsection{Results and Analysis}
\subsubsection{Comparison with self-supervised methods}

The experimental results are shown in Table \ref{t2}. Overall, our self-supervised learning framework can significantly surpass the previous state-of-the-art methods. Especially in LSST and JapaneseVowels whose time lengths are relative short, methods based on fitting specific values or features like TST, TS-TCC and NVP cannot perform well, while our framework can obtain promising and stable performances. It demonstrates that simple trend predicting is more efficient than regression. W2V and our framework are suitable for both short and long time length datasets, and our performances can significantly surpass W2V. This demonstrates the powerful representation learning ability of our framework. Moreover, our self-supervised framework is shown to be superior to the supervised way in two of four datasets while in the other two datasets can achieve similar results. It also proves that our self-supervised learning framework is capable of learning not only complex characteristics between samples but also within samples.

For a more convincing comparison, we conduct experiments by replacing the origin encoder of the previous methods with our proposed encoder. It helps to offer a more detailed view on the aspects of both encoder and pretext task. It is shown in Table \ref{t2} that applying our encoder like TS-TCC+ and TST+ helps to improve their effectiveness, which demonstrates the encoding ability of our encoder. W2V is a method which pays more attention on the local information, so the causal dilated convolutions which only focuses on previous information is more suitable than W2V+ which encodes the global information. In the aspect of pretext task, when combining NTP task and CS task with our encoder, the self-supervised learning ability is further improved by a large margin compared to other pretext tasks.

\begin{figure*}
\centering
\includegraphics[width=1\textwidth]{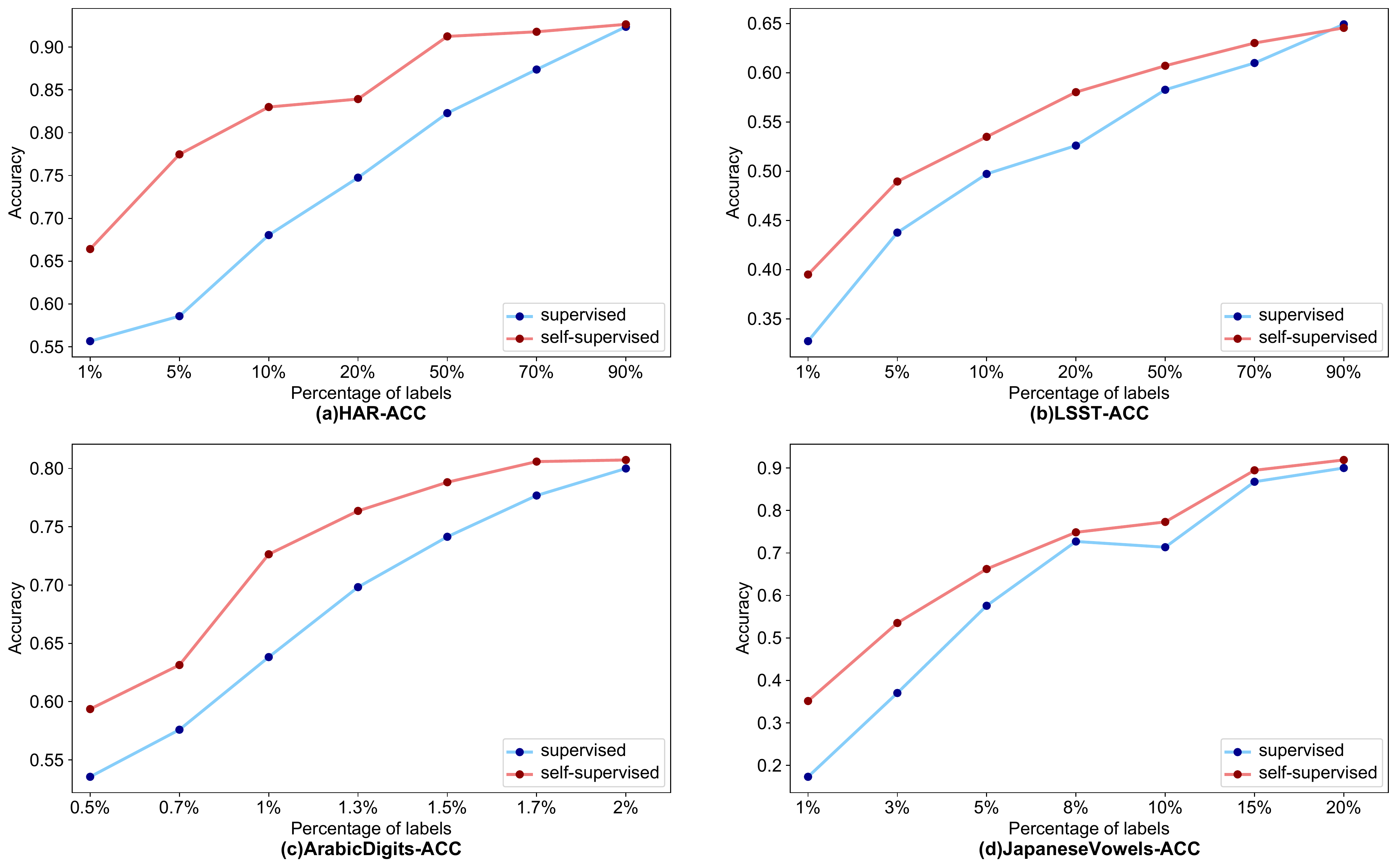} % Reduce the figure size so that it is slightly narrower than the column.
\caption{Few-shot learning results. We report the comparison between our self-supervised framework and the supervised way.}
\label{fig3}
\end{figure*}

\subsubsection{Few-shot learning}

To further prove the effect of our self-supervised framework, we conduct few-shot learning experiments comparing with the supervised learning. In HAR and LSST datasets, we choose 1\%, 5\%, 10\%, 20\%, 50\%, 70\%, 90\% percentage of labeled samples for model training respectively. For ArabicDigits and JapaneseVowels datasets, we adopt a set of smaller percentages in order to compare the performance of few-shot learning more clearly. The results are shown in Figure \ref{fig3}. Among these datasets, our self-supervised framework can significantly surpass the supervised learning by training a single fully connected layer with limited labeled samples.

\subsection{Ablation Study}
\subsubsection{Ablation study on pretext task}

To analyze the role of each of our pretext tasks, we apply the following variants as comparisons: \\
\textbf{(1) -NTP}: It only applies Contextual Similarity task to self-supervised learning. \\
\textbf{(2) -CS}: It only applies Next Trend Prediction task to self-supervised learning. \\
\textbf{(3) -neg augment}: It removes the negative samples generated by the asynchronous disorder strategy. \\
\textbf{(4) reverse neg}: It regards the samples generated by the asynchronous disorder strategy as positive samples.

The experimental results are shown in Table \ref{t3}. It can be seen that in our pretext tasks, NTP task and CS task can well cooperate with each other. Specifically, NTP task occupies the most important position in small datasets while CS task is more important in datasets with large number of samples. On the one hand, it shows the importance of the internal relationships between channels to multivariate time series. On the other hand, it demonstrates that self-supervised learning should not only focus on the characteristics of the sample itself, but also need to maintain the independence from other samples. In addition, negative sample enhancement can bring a more stable effect to the encoder and further enhance the effect of self-supervised learning method. The conversion of the negative samples into positive samples will lead to the decline of the effect, which shows that in the time series, the time relationships between channels can not be disturbed.

\begin{table*}
\centering
\caption{Ablation study on pretext task.}
\scalebox{0.67}{
\begin{tabular}{lcccccccc}
\specialrule{0em}{0.5pt}{0.5pt}
\toprule
& \multicolumn{2}{c}{\textbf{HAR}} & \multicolumn{2}{c}{\textbf{LSST}} & \multicolumn{2}{c}{\textbf{ArabicDigits}} & \multicolumn{2}{c}{\textbf{JapaneseVowels}} \\ \midrule
\textbf{Method} & \textbf{ACC} & \textbf{MF1} & \textbf{ACC} & \textbf{MF1} & \textbf{ACC} & \textbf{MF1} & \textbf{ACC} & \textbf{MF1}\\   \midrule

\textbf{-NTP} & 90.96$\pm$0.59 & 90.71$\pm$0.59 & 61.05$\pm$1.20 & 41.30$\pm$0.40 & 95.55$\pm$0.36 & 95.50$\pm$0.41 & 94.87$\pm$0.27 & 94.87$\pm$0.22  \\  
\textbf{-CS} & 76.94$\pm$0.66 & 75.96$\pm$0.67 & 53.00$\pm$0.81 & 33.50$\pm$1.97 & 90.71$\pm$0.06 & 90.72$\pm$0.08 & 97.71$\pm$0.13 & 97.69$\pm$0.13  \\  
\textbf{-neg augment} & 87.06$\pm$0.25 & 86.79$\pm$0.23 & 61.88$\pm$0.57 & 45.21$\pm$0.10 & 96.32$\pm$0.32 & 96.32$\pm$0.32 & 97.98$\pm$0.13 & 97.79$\pm$0.16  \\  
\textbf{reverse neg} & 83.02$\pm$0.22 & 82.65$\pm$0.15 & 63.30$\pm$0.65 & 45.40$\pm$0.50 & 92.62$\pm$0.43 & 92.61$\pm$0.44 & 97.42$\pm$0.16 & 97.22$\pm$0.09  \\  
\textbf{CaSS} & \textbf{92.57$\pm$0.24} & \textbf{92.40$\pm$0.17} & \textbf{64.94$\pm$0.02} & \textbf{46.11$\pm$0.55} & \textbf{97.07$\pm$0.20} & \textbf{97.07$\pm$0.20} & \textbf{98.11$\pm$0.27} & \textbf{98.14$\pm$0.08}  \\  

\bottomrule
\end{tabular}}

\label{t3}
\end{table*}

\subsubsection{Ablation study on encoder}

To analyze the effect of each component in the encoder, we apply the following variants for comparisons by supervised learning: \\ 
\textbf{(1) Self Aggregate}: It contains two Transformers with self-attention mechanism to encode time-wise and channel-wise features independently. Finally the time-wise features are fused into channel-wise features through Aggregate Layer. \\ 
\textbf{(2) Channel Self}: A single Transformer is applied with self-attention mechanism to encode channel-wise features. \\ 
\textbf{(3) -Aggregate Layer}: The channel-wise features of the last Co-Transformer Layer are applied without fusing the time-wise features.

The experimental results are shown in Table \ref{t4}. It can be seen that, if time-wise and channel-wise features are only fused in the last aggregate layer without interactions in the previous stage, they cannot be well integrated and bring the loss of information. As contrast, the interactions between the two Transformers can significantly bring the performance improvement. The results of Channel Self illustrate the importance of each channel's independent time pattern. Finally, the existence of Aggregate Layer can also better integrate the features of time-wise and channel-wise while alleviating the redundancy of features.

\begin{table*}
\centering
\caption{Ablation study on encoder.}
\scalebox{0.67}{
\begin{tabular}{lcccccccc}
\specialrule{0em}{0.5pt}{0.5pt}
\toprule
& \multicolumn{2}{c}{\textbf{HAR}} & \multicolumn{2}{c}{\textbf{LSST}} & \multicolumn{2}{c}{\textbf{ArabicDigits}} & \multicolumn{2}{c}{\textbf{JapaneseVowels}} \\ \midrule
\textbf{Method} & \textbf{ACC} & \textbf{MF1} & \textbf{ACC} & \textbf{MF1} & \textbf{ACC} & \textbf{MF1} & \textbf{ACC} & \textbf{MF1}\\   \midrule

\textbf{Self Aggregate} & 91.88$\pm$0.35 & 91.85$\pm$0.38 & 66.10$\pm$0.24 & 42.80$\pm$0.83 & 96.64$\pm$0.22 & 96.63$\pm$0.23 & 97.17$\pm$0.13 & 96.94$\pm$0.10  \\  
\textbf{Channel Self} & \textbf{93.15$\pm$0.91} & \textbf{93.13$\pm$0.95} & 56.45$\pm$0.48 & 31.31$\pm$0.50 & 97.48$\pm$0.11 & 97.47$\pm$0.12 & 97.57$\pm$0.27 & 97.42$\pm$0.22  \\  
\textbf{-Aggregate Layer} & 92.35$\pm$0.12 & 92.36$\pm$0.13 & 63.97$\pm$0.59 & 42.65$\pm$0.18 & 97.68$\pm$0.32 & 97.68$\pm$0.32 & 97.57$\pm$0.27 & 97.38$\pm$0.23  \\  
\textbf{CaT} & 92.35$\pm$0.63 & 92.40$\pm$0.58 & \textbf{66.57$\pm$0.38} & \textbf{51.60$\pm$1.26} & \textbf{98.07$\pm$0.38} & \textbf{98.07$\pm$0.38} & \textbf{97.71$\pm$0.13} & \textbf{97.56$\pm$0.07}  \\  

\bottomrule
\end{tabular}}

\label{t4}

\end{table*}

\section{Conclusion}

Self-supervised learning is essential for multivariate time series. In this work, we propose a new self-supervised learning framework CaSS to learn the complex representations of MTS. For the encoder, we propose a new Transformer-based encoder Channel-aware Transformer to capture the time-wise and channel-wise features more efficiently. For the pretext task, we propose Next Trend Prediction from the perspective of channel-wise for the first time and combine it with Contextual Similarity task. These novel pretext tasks can well cooperate with our encoder to learn the  characteristics. Our self-supervised learning framework demonstrates significant improvement on MTS classification comparing with previous works, and can significantly surpass supervised learning with limited labeled samples.

\section{Acknowledgments}
This work was supported by the National Key Research and Development Program of China, No.2018YFB1402600.

\bibliographystyle{splncs04}
\bibliography{CaSS}

\end{document}